\documentclass[letterpaper, 10 pt, conference]{ieeeconf}  

\IEEEoverridecommandlockouts                              

\usepackage{bm}
\usepackage{amsmath}
\usepackage{amssymb}
\usepackage{graphicx}
\usepackage[ruled,vlined,linesnumbered]{algorithm2e}
\usepackage{algpseudocode}
\usepackage{arydshln}
\usepackage{booktabs}
\usepackage{overpic}
\usepackage{tikz}
\usepackage{circledsteps}
\usepackage{todonotes}

\makeatletter
\let\MYcaption\@makecaption
\makeatother

\usepackage[subrefformat=parens]{subcaption}
\makeatletter
\let\@makecaption\MYcaption
\makeatother

\title{\LARGE \bf
COLSON: Controllable Learning-Based Social Navigation \\ via Diffusion-Based Reinforcement Learning 
}

\author{Kohei Matsumoto$^{1}$$^{\dag}$, Yuki Tomita$^{2}$$^{\dag}$, Yuki Hyodo$^{2}$, and Ryo Kurazume$^{1}$
\thanks{*This work was not supported by any organization}
\thanks{$^{1}$Kohei Matsumoto and Ryo Kurazume are with the Faculty of Information Science and Electrical Engineering, Kyushu University, Fukuoka 819-0395, Japan {\tt \small matsumoto@ait.kyushu-u.ac.jp}, {\tt \small kurazume@ait.kyushu-u.ac.jp}}%
\thanks{$^{2}$Yuki Tomita and Yuki Hyodo are with the Graduate School of Information Science and Electrical Engineering, Kyushu University, Fukuoka 819-0395, Japan
        {\tt\small tomita@irvs.ait.kyushu-u.ac.jp}, {\tt\small hyodo@irvs.ait.kyushu-u.ac.jp}}%
\thanks{$^{\dag}$Authors contributed equally}%
}

\begin{document}

\maketitle
\thispagestyle{empty}
\pagestyle{empty}

\begin{abstract}
Mobile robot navigation in dynamic environments with pedestrian traffic is a key challenge in the development of autonomous mobile service robots. Recently, deep reinforcement learning-based methods have been actively studied and have outperformed traditional rule-based approaches owing to their optimization capabilities. 
Among these methods, those that assume continuous action spaces typically rely on Gaussian distributions, which limit the flexibility of the generated actions.
In contrast, the application of diffusion models to reinforcement learning has advanced, enabling more flexible action distributions than Gaussian policy-based approaches. In this study, we apply a diffusion-based reinforcement learning approach to social navigation and validate its effectiveness.
Furthermore, by exploiting the characteristics of diffusion models, we propose extensions that enable adaptation to previously unseen scenarios without additional training.
As concrete scenario examples, we demonstrate adaptability to scenarios in which static obstacles exist in the environment that were not present during training, as well as scenarios in which the objective differs from training, such as accompanying target pedestrians while avoiding others to reach the destination. 
\end{abstract}
\section{INTRODUCTION}
Social navigation, which enables safe and efficient navigation in dynamic environments with pedestrian traffic, is crucial for realizing autonomous mobile service robots. In recent years, deep reinforcement learning-based methods have been extensively studied owing to advancements in machine learning.
Among these, methods that assume continuous action spaces require modeling of the action distribution using predefined functions, such as Gaussian distributions.
Generative models have advanced rapidly in recent years, with diffusion models achieving remarkable success in tasks such as image generation. Their application to reinforcement learning has also been explored, demonstrating strong performance in continuous control tasks due to their expressive capacity and lack of reliance on Gaussian distribution assumptions. 
Motivated by this, we apply a diffusion-based reinforcement learning approach to mobile robot navigation in dynamic environments with pedestrian traffic.
This method demonstrates superior performance compared with Gaussian policy-based approaches, maintaining high performance even when the number of pedestrians in the environment or their behavioral patterns change.

\par
\begin{figure}[t]
 \centering
  \includegraphics[width=80mm]{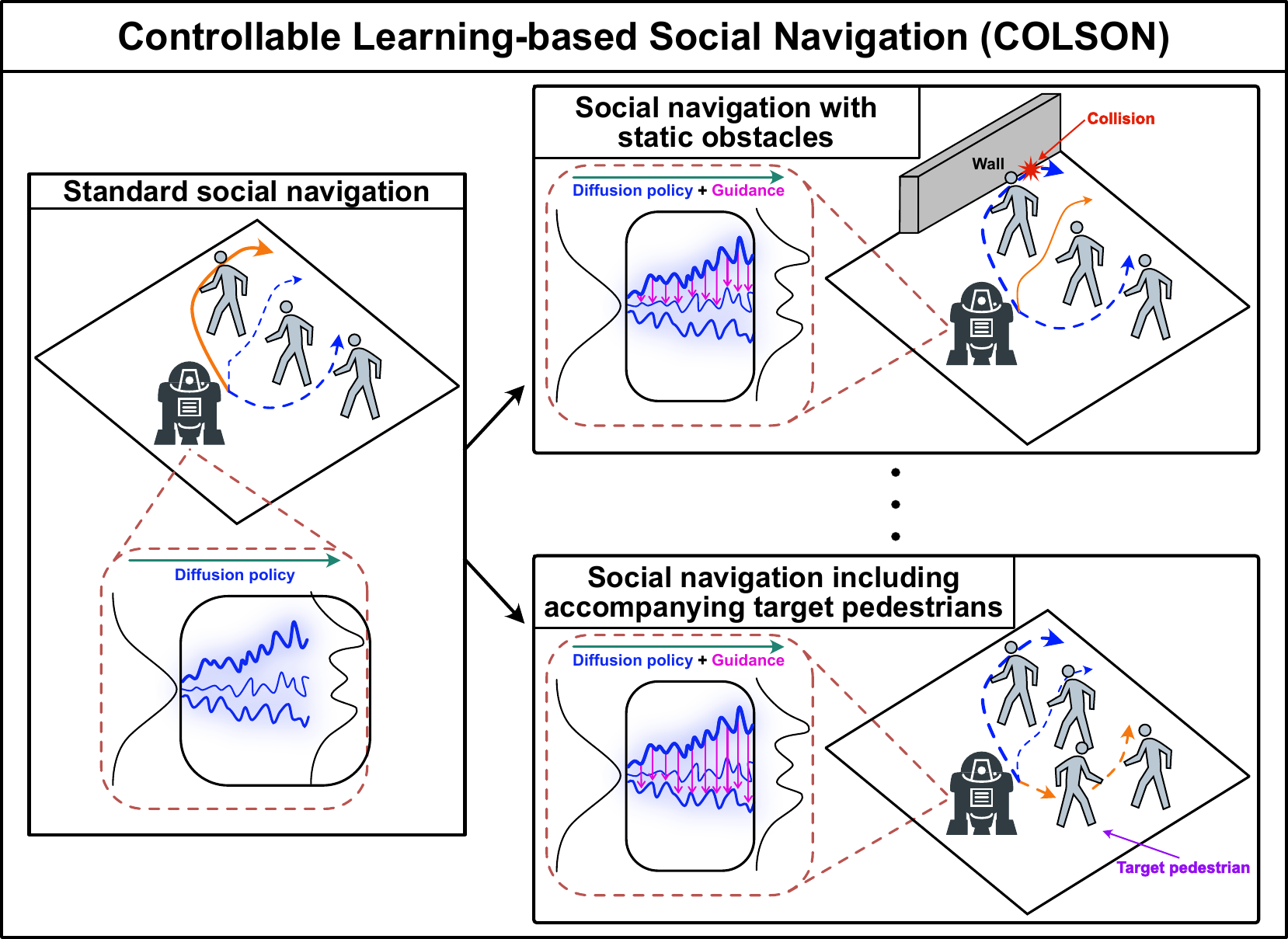}
   \caption{Conceptual diagram of the proposed method. In standard social navigation, the proposed method leverages the multimodality of diffusion models to generate various actions that allow robots to avoid pedestrians. The blue dashed arrows represent examples of candidate actions generated in this situation, whereas the solid orange line indicates the actual action selected. In environments with static obstacles, actions generated solely by the diffusion model may result in collisions with walls. Therefore, guidance is applied to steer the selection toward alternative candidates that avoid wall collisions. In tasks involving pedestrian following, diffusion models alone tend to generate behaviors that move away from pedestrians. Therefore, guidance is applied to generate actions that follow the target pedestrian.}
  \label{fig:conceptual_diagram}
  \vspace{-2mm}
\end{figure}
\textcolor{black}{
Furthermore, the proposed method leverages the diverse action generation capability of policies obtained through diffusion-based reinforcement learning to adapt to environments and tasks not considered during training without additional training.
We consider two concrete scenarios. The first scenario pertains to social navigation in environments with static obstacles that are not present during the training phase. The second scenario pertains to the companion task. This objective is not considered during training. In this task, agents are required to follow a designated pedestrian while avoiding other pedestrians to achieve their objectives.
}
Considering these adaptability features, the proposed method is called \textbf{Co}ntrollable \textbf{L}earning-based \textbf{So}cial \textbf{N}avigation (COLSON). A conceptual diagram of COLSON is shown in Fig. \ref{fig:conceptual_diagram}.\par

The main contributions of this study are as follows.
\begin{itemize}
    \item We integrate a diffusion-based reinforcement learning method with a graph neural network (GNN) architecture and apply it to social navigation tasks. To the best of our knowledge, this is the first application of diffusion-based reinforcement learning to social navigation.
    \item We propose an annealing method to improve the performance of diffusion-based reinforcement learning for social navigation.
    \item  We demonstrate that the proposed method can handle conditions that are not considered during training, such as adaptation to environments with static obstacles and companion tasks, without requiring additional training.
    \item  Through comparative experiments with conventional methods, we show that the proposed method outperforms existing approaches and exhibits high performance, even in environments where the number of pedestrians differs from that in the training phase, highlighting its scalability.
    \item We conduct a real-world demonstration to validate that the proposed method can be applied to an actual robot in practical situations.
\end{itemize}

\section{RELATED WORK}
\subsection{Social Navigation}
Numerous deep reinforcement learning-based social navigation methods have been proposed. Chen et al. \cite{Chen2017-ni} introduced a value-based learning model for multi-agent collision avoidance, marking a pioneering achievement in deep reinforcement learning approaches aimed at enabling autonomous mobile robots to safely navigate dynamic environments. Building on this, Chen et al. \cite{Chen2017-ru} proposed a method that incorporates social norms into the reward function to improve pedestrian collision avoidance and further addressed the challenge of pedestrian switching in multi-agent scenarios by sharing weights in the pedestrian information processing layer and employing max pooling. In addition to the direct application of deep reinforcement learning, architectural improvements in neural networks have been explored to enhance performance and develop methods that can accommodate an arbitrary number of agents \cite{Everett2018-cq, Chen2019-ju, Chen2020-rj, MatsumotoSII2022, Yang2023-yy, Liu2023-tc}. Other studies have expanded the scope beyond performance improvement and scalability to pedestrian density by investigating approaches that leverage human gaze information \cite{Chen2020-fp}, utilizing map-based processing to enable navigation in environments with both static and dynamic obstacles \cite{Liu2020-wn, Yao2021-vq}, and incorporating continuous action spaces \cite{Zhang2021-uy}. More recently, hybrid approaches that integrate reinforcement learning-based methods with rule-based methods have been proposed to further enhance robustness and adaptability \cite{Wu2023-xf, Matsumoto2024-dd}.

Despite extensive research on social navigation, no studies have employed diffusion-based reinforcement learning approaches to generate both high-performance and diverse behaviors that can be adaptively guided according to specific situations or that can be adapted to novel environments and emerging applications.
\subsection{Diffusion Models for Action Generation} 
Diffusion models have achieved remarkable success in various data generation tasks, including image synthesis \cite{Ho2020-ls, Rombach2022-tb, Nakashima2024-gt}. Recent studies have extended their applications beyond vision-based data to action generation in robotics through imitation learning and reinforcement learning. 
In the field of imitation learning, methods based on behavior cloning (BC) have been proposed \cite{Janner2022-mj, Reuss2023-qp, Chi2023-nd}, where diffusion models are trained to generate actions that replicate target demonstration data, making it the most fundamental approach for applying diffusion models to action generation. 
In addition to imitation learning, diffusion models have been applied to offline reinforcement learning \cite{Wang2023-gn, Terry-Suh2023-cv, Kang2023-pd, Hansen-Estruch2023-gy}, where exploration is not required, allowing models to learn distributions from datasets in a manner similar to imitation learning.  
More recently, efforts have been directed toward applying diffusion models to online reinforcement learning \cite{Yang2023-te}, particularly in settings that differ from conventional diffusion model formulations, such as learning without access to presampled data from an optimal target distribution \cite{Psenka2024-mf} and approaches based on weighted regression \cite{Ding2024-dq}. 

Furthermore, in the context of applying diffusion models to mobile robots, methods leveraging diffusion policy for goal-conditioned navigation and undirected exploration have been proposed \cite{Sridhar2024-uo}. Other approaches use models trained on ground truth data from the A$^{\star}$ algorithm for global path planning \cite{Liu2024-jv, Stamatopoulou2024-my}. \par

From the perspective of guidance of diffusion models,  methods capable of generating actions that simultaneously satisfy multiple constraints or combine multiple skills \cite{Ajay2023-cq}, controllable learning-based pedestrian simulation \cite{Rempe2023-rj}, and a visual navigation method that utilizes depth estimation-based cost guidance have been proposed, demonstrating high performance even in unseen environments \cite{Zeng2025-mx}.

\textcolor{black}{
However, the application of diffusion-based reinforcement learning to social navigation remains largely unexplored.
}

\section{PRELIMINARIES}
This study addresses the problem of two-dimensional (2D) mobile robot navigation while ensuring collision avoidance with pedestrians.
The state of the robot includes its position from the goal, velocity, and orientation information: $\boldsymbol{s}^r = \left[^{g}\boldsymbol{p}^{r}, \boldsymbol{v}^{r}, \theta\right]$.
In contrast, the position and velocity information of each pedestrian is converted to a robot-centered coordinate system and adopted as a state $\boldsymbol{s}^n = \left[^{rc}\boldsymbol{p}^n, ^{rc}\boldsymbol{v}^n\right]$. The set of pedestrian states in the environment is denoted as $\boldsymbol{s}^h = \left[\boldsymbol{s}^1,\boldsymbol{s}^2,\dots,\boldsymbol{s}^n\right]$.
The robot’s action includes velocity values in the X- and Y-directions, $\boldsymbol{a}=\left[v^{c}_{x}, v^{c}_{y}\right]$. 
The robot obtains rewards after performing actions. \textcolor{black}{In this study, we adopt the form of the reward function from previous research \cite{Chen2019-ju}.} $R_t$ denotes the reward function at time $t$ as follows:
\begin{align}
    \label{eq:reward}
    R_t=\left\{\begin{array}{ll}
    -0.25 & \text { if } d_{t}<0 \\
    -0.1+d_{t} / 2 & \text { else if } d_{t}<0.2 \\
    1 & \text { else if } \boldsymbol{p}^{r}_{t}=\boldsymbol{p}^{g}_{t} \\
    0 & \text { otherwise }
    \end{array}\right.,
\end{align}
where $d_{t}$ denotes the minimum separation distance between the robot and pedestrians, $\boldsymbol{p}_{t}^{r}$ denotes the robot's position, and $\boldsymbol{p}_{t}^{g}$ denotes the goal position at time $t$.
\section{APPROACH}

This section presents the model architecture, training methodology, and guidance methods for static obstacle avoidance and companion tasks.

\subsection{Model Architecture}
\begin{figure}[ht]
 \centering
  \includegraphics[width=80mm]{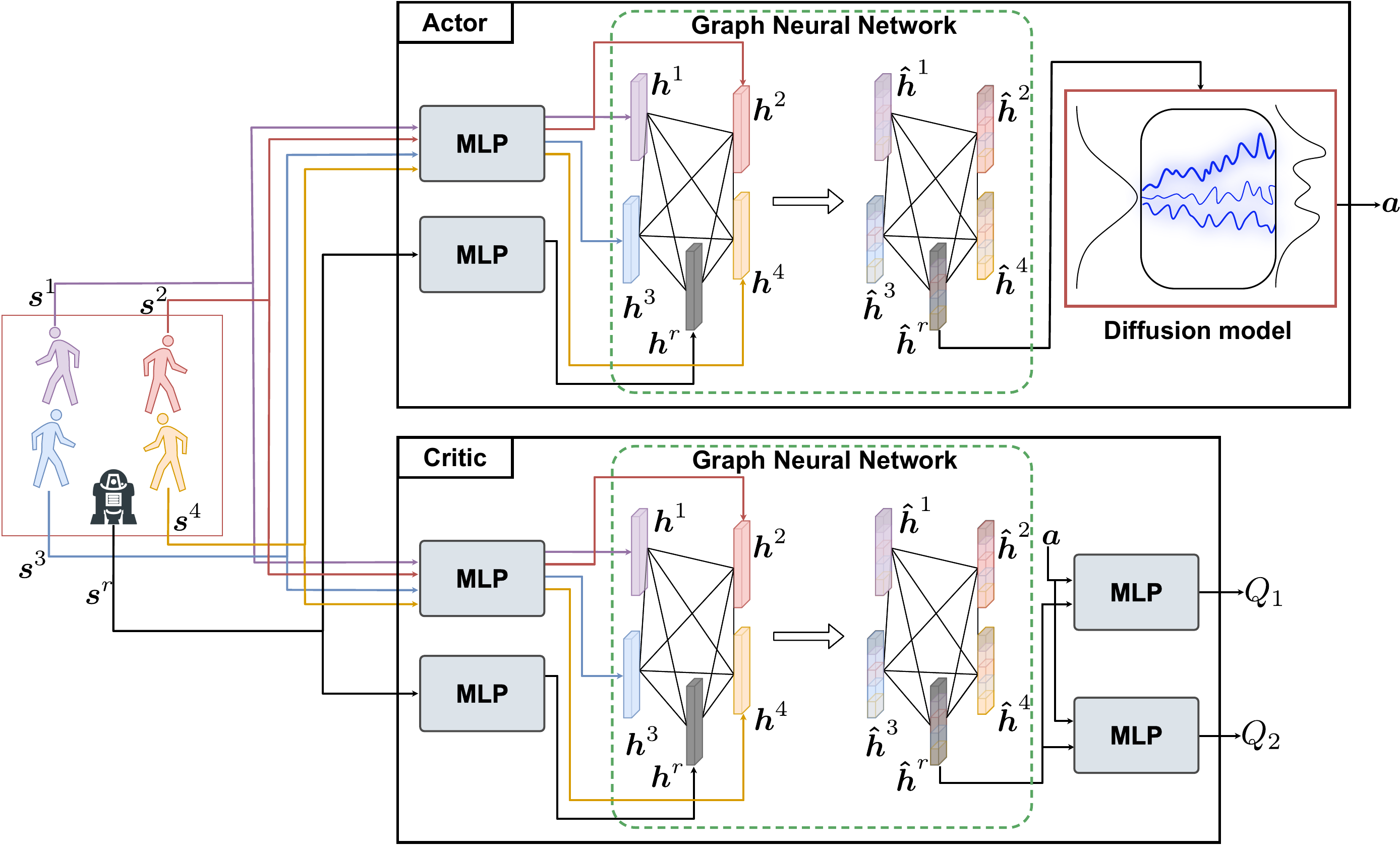}
  \caption{Architecture of proposed method. $\boldsymbol{s}^r$ and $\boldsymbol{s}^n$ indicate the states of each robot and pedestrian, respectively. $\boldsymbol{h}^n$ indicates the features of each robot and pedestrian. $\boldsymbol{\hat{h}}^n$ indicates the features after the GNN is applied.}
  \label{fig:architecture}
\end{figure}

The model architecture of the proposed method is shown in Fig. \ref{fig:architecture}.
This model adopts an actor–critic framework in which both the actor and critic incorporate GNNs. 
Each GNN utilizes an attention-based mechanism, as utilized in a related study \cite{Chen2020-rj}.
The actor employs a diffusion model that considers the output of the GNN as a condition and generates actions. The critic network inputs the output of the GNN into a multilayer perceptron to compute Q-values.
\subsection{Q-Score Matching and Annealing}
\begin{algorithm}[b]
    \SetAlgoLined
    \DontPrintSemicolon
    \caption{Training algorithm using QSM}
    \label{alg:qsm}
    {\footnotesize
    Initialize the score network $\Psi_{\theta}$ and critic networks $Q_{\phi^1}$ and $Q_{\phi^2}$\;
    Set the parameter values of the target critics $Q_{\phi^{\prime, 1}}$ and $Q_{\phi^{\prime, 2}}$ equal to those of the main critics \;
    \For{$i=1$ \KwTo $E$}{
        Explore using the policy until finishing an episode\\
        After finishing an episode, store the trajectory of $\left(\boldsymbol{s}_{t}, \boldsymbol{a}_{t}, r_{t}, \boldsymbol{s}^{\prime}_{t}\right)$ to the replay buffer $\mathcal{D}$\\
        \BlankLine
        \BlankLine
        \BlankLine
        Sample batch $\mathcal{B} = \{(\boldsymbol{s}, \boldsymbol{a}, r, \boldsymbol{s}^{\prime})$\} from the replay buffer $\mathcal{D}$\;
        Sample actions for computing targets
        $\quad \boldsymbol{\hat{a}}^{\prime} \sim \pi_\theta\left(\cdot \mid \boldsymbol{s}^{\prime}\right)$\;
        Calculate targets for the Q-function
        $y\left(r, \boldsymbol{s}^{\prime}\right)=r+\gamma\left(\min _{i=1,2} Q_{\phi^{\prime, i}}\left(\boldsymbol{s}^{\prime}, \boldsymbol{\hat{a}}^{\prime}\right)\right)$\\
        Update the critics by minimizing $L_{\text{critic}} = \displaystyle \frac{1}{|\mathcal{B}|} \sum(Q_{\phi^i}(\boldsymbol{s}, \boldsymbol{a}) -  y\left(r, \boldsymbol{s}^{\prime}\right))^2$ \hspace{0.1cm} for $i = 1,2$\;
        Update the score network by minimizing $L_{\text{QSM}} = \displaystyle \frac{1}{|\mathcal{B}|} \sum \lVert \Psi_{\theta}(\boldsymbol{s}, \boldsymbol{a})-\alpha \nabla_{a} Q(\boldsymbol{s}, \boldsymbol{a}) \rVert^2$\;
        Update the target critic using polyak averaging $\phi^{\prime,i} \leftarrow \rho \phi^{\prime,i}+(1-\rho) \phi_{i}$ \hspace{0.1cm} for $i = 1,2$. \\;
    }
    }
\end{algorithm}
In this study, we use Q-score matching (QSM) \cite{Psenka2024-mf} as the training framework. Diffusion models are typically trained using score functions for the target distribution. However, in reinforcement learning, where the objective is to learn an optimal policy, the optimal policy is unknown. Consequently, neither the score function nor samples from the optimal policy can be precomputed.
To overcome this challenge, QSM facilitates reinforcement learning using diffusion models by leveraging the action gradient of the Q-function as a score.
The training procedure using QSM is presented in Algorithm \ref{alg:qsm}. 
\textcolor{black}{
The coefficient $\alpha$ represents the inverse temperature parameter.
Although smaller values of $\alpha$ enhance exploration during reinforcement learning, they also make it more difficult for the final action distribution to converge. Conversely, larger values of $\alpha$ produce the opposite effect. While the previous study has fixed this coefficient during training, we propose an annealing technique that gradually varies $\alpha$ during training. 
In this study, we adopt the value that achieved the highest return from our preliminary experiment as the target value $\hat{\alpha}$. 
The proposed method linearly anneals $\alpha$ from 1 to the target value $\hat{\alpha}$ during training. This schedule allows the agent to explore broadly in the early stages of training while promoting convergence toward optimal actions in later stages.
}
\subsection{\textcolor{black}{Guidance for Static Obstacle Avoidance}}
In this section, we describe the first example of guidance for action generation to realize social navigation with static obstacles.
The training environment in this study contains only pedestrians, implying that the trained policy lacks the capability to avoid static obstacles. 
\begin{table*}[ht]
    \vspace{-0.20cm}
    \centering
    \caption{Numerical comparison in circle crossing scenario with 5 ORCA pedestrians}
    \vspace{-1mm}
    \label{tab:orca_circle}
    \scalebox{0.80}{
    \begin{tabular}{lccccccccc}
        \toprule
        \multicolumn{1}{c}{} & 
        \multicolumn{4}{c}{Visible} & 
        \multicolumn{4}{c}{Invisible} \\ 
        \cline{2-5} \cline{7-10}
        \\
        Method
        & Success [\%]   $\uparrow$       
        & Collision [\%] $\downarrow$        
        & Exec. time [s] $\downarrow$ 
        & Return         $\uparrow$ 
        &
        & Success [\%]   $\uparrow$       
        & Collision [\%] $\downarrow$        
        & Exec. time [s] $\downarrow$ 
        & Return         $\uparrow$ 
        \\ \hline
        ORCA (for reference)
        & $ 100.00 \pm 0.00 $
        & $ 0.00  \pm 0.00 $
        & $ 10.02 \pm 0.00 $
        & $ 0.542 \pm 0.000 $
        &
        & $ 42.80 \pm 0.00 $
        & $ 56.80  \pm 0.00 $
        & $ 10.93 \pm 0.00 $
        & $ 0.081 \pm 0.000 $
        \\ \hdashline
        BC
        & $ 77.28 \pm 0.17 $
        & $ 22.56 \pm 0.18 $
        & $ 10.06 \pm 0.61 $
        & $ 0.377 \pm 0.001 $
        &
        & $ 5.08 \pm 0.21 $
        & $ 94.92 \pm 0.21 $
        & $ 10.24 \pm 0.81 $
        & $ -0.182 \pm 0.002 $
        \\
        AWR  
        & $ 87.60 \pm 1.96 $
        & $ 12.36 \pm 1.97 $
        & $ 11.44 \pm 1.80 $
        & $ 0.466 \pm 0.009 $
        &
        & $ 42.44 \pm 4.78 $
        & $ 57.56 \pm 4.78 $
        & $ 11.48 \pm 2.16 $
        & $ 0.122 \pm 0.030 $
        \\
        CAWR 
        & $ 61.00 \pm 0.57 $
        & $ 35.28 \pm 0.45 $
        & $ 11.98 \pm 8.13 $
        & $ 0.236 \pm 0.001 $
        &
        & $ 12.48 \pm 2.11 $
        & $ 86.68 \pm 2.34 $
        & $ 12.96 \pm 10.21 $
        & $ 0.122 \pm 0.030$
        \\
        QVWR  
        & $ 62.48 \pm 3.84 $
        & $ 34.24 \pm 3.86 $
        & $ 11.44 \pm 5.63 $
        & $ 0.245 \pm 0.020 $
        &
        & $ 6.68 \pm 1.00 $
        & $ 92.92 \pm 1.10 $
        & $ 11.44 \pm 6.04 $
        & $ -0.173 \pm 0.007$
        \\
        AWAC 
        & $ 71.08 \pm 2.91 $
        & $ 28.92 \pm 2.91 $
        & $ 11.32 \pm 5.83 $
        & $ 0.327 \pm 0.011 $
        &
        & $ 29.76 \pm 7.20 $
        & $ 70.24 \pm 7.20 $
        & $ 11.31 \pm 5.69 $
        & $ 0.010 \pm 0.038 $
        \\
        DIPO
        & $ 80.80 \pm 1.97 $
        & $ 17.52 \pm 1.42 $
        & $ 11.29 \pm 1.26 $
        & $ 0.327 \pm 0.011$
        &
        & $ 30.32 \pm 3.41 $
        & $ 68.96 \pm 3.03 $
        & $ 12.22 \pm 0.15 $
        & $ 0.022 \pm 0.021 $
        \\
        QVPO
        & $ 98.24 \pm 0.02 $
        & $ 1.68 \pm 0.01 $
        & $ 9.75 \pm 0.97 $
        & $ 0.618 \pm 0.001 $
        &
        & $ 97.44 \pm 0.03 $
        & $ 2.36 \pm 0.03 $
        & $ 9.99 \pm 0.79 $
        & $ 0.606 \pm 0.001 $
        \\
        QSM
        & $ 98.28 \pm 0.03 $
        & $ 1.60 \pm 0.03 $
        & $ \mathbf{8.70} \pm \mathbf{0.07} $
        & $ \mathbf{0.647} \pm \mathbf{0.000} $
        &
        & $ 98.44 \pm 0.01 $
        & $ 1.32 \pm 0.01 $
        & $ \mathbf{9.05} \pm \mathbf{0.05} $
        & $\mathbf{0.640} \pm \mathbf{0.000} $
        \\
        QSM-A (w/ annealing)
        & $ \mathbf{98.68} \pm \mathbf{0.00} $ 
        & $ \mathbf{0.84} \pm \mathbf{0.00} $ 
        & $ 9.24 \pm 0.23 $ 
        & $ 0.637 \pm 0.000$  
        & & $ \mathbf{98.88} \pm \mathbf{0.00} $ 
        & $ \mathbf{0.60} \pm \mathbf{0.00} $ 
        & $ 9.37 \pm 0.20 $ 
        & $ 0.636 \pm 0.000$ 
        \\ 
        \bottomrule
    \end{tabular}
    }
\end{table*}\par
\begin{table*}[ht]
    \centering
    \caption{Numerical comparison in circle crossing scenario with 5 social force pedestrians}
    \vspace{-1mm}
    \label{tab:socialforce_circle}
    \scalebox{0.80}{
    \begin{tabular}{lccccccccc}
        \toprule
        \multicolumn{1}{c}{} & 
        \multicolumn{4}{c}{Visible} & 
        \multicolumn{4}{c}{Invisible} \\ 
        \cline{2-5} \cline{7-10}
        \\
        Method
        & Success [\%]   $\uparrow$       
        & Collision [\%] $\downarrow$        
        & Exec. time [s] $\downarrow$ 
        & Return         $\uparrow$
        & 
        & Success [\%]   $\uparrow$       
        & Collision [\%] $\downarrow$        
        & Exec. time [s] $\downarrow$ 
        & Return         $\uparrow$
        \\ \hline
        ORCA (for reference)
        & $ 90.80 \pm 0.00 $
        & $ 9.20  \pm 0.00 $
        & $ 9.79 \pm 0.00 $
        & $ 0.544 \pm 0.000 $
        & 
        & $ 19.00 \pm 0.00 $
        & $ 81.00  \pm 0.00 $
        & $ 10.82 \pm 0.00 $
        & $ -0.088 \pm 0.000 $
        \\ \hdashline
        BC
        & $ 63.4 \pm 1.30 $ 
        & $ 36.6 \pm 1.30 $ 
        & $ 10.16 \pm 0.73 $ 
        & $ 0.303 \pm 0.007$
        & 
        & $ 0.28 \pm 0.00 $
        & $ 99.72 \pm 0.00 $
        & $ 14.62 \pm 75.05 $
        & $ -0.222 \pm 0.000$
        \\
        AWR  
        & $ 86.92 \pm 0.57 $
        & $ 13.04 \pm 0.58 $
        & $ 11.52 \pm 2.18 $
        & $ 0.483 \pm 0.007 $
        & 
        & $ 41.68 \pm 11.13 $
        & $ 58.28 \pm 11.16 $
        & $ 11.37 \pm 2.00 $
        & $ 0.131 \pm 0.076 $
        \\
        CAWR 
        & $ 63.32 \pm 1.94 $
        & $ 34.20 \pm 2.25 $
        & $ 12.28 \pm 11.09 $
        & $ 0.278 \pm 0.005 $
        &
        & $ 6.80 \pm 0.90 $
        & $ 92.48 \pm 1.12 $
        & $ 21.17 \pm 70.54 $
        & $ -0.172 \pm 0.006 $
        \\
        QVWR  
        & $ 58.04 \pm 0.89 $
        & $ 36.36 \pm 1.91 $
        & $ 12.46 \pm 7.28 $
        & $ 0.245 \pm 0.003 $
        & 
        & $ 3.60 \pm 0.33 $
        & $ 96.08 \pm 0.38 $
        & $ 23.85 \pm 70.94 $
        & $ -0.196 \pm 0.003 $
        \\
        AWAC 
        & $ 80.64 \pm 3.24 $
        & $ 19.32 \pm 3.25 $
        & $ 11.47 \pm 7.14 $
        & $ 0.431 \pm 0.022 $
        & 
        & $ 42.68 \pm 6.67 $
        & $ 57.32 \pm 6.67 $
        & $ 15.53 \pm 70.50 $
        & $ 0.127 \pm 0.048 $
        \\
        DIPO
        & $ 73.52 \pm 0.41 $
        & $ 25.00 \pm 0.24 $
        & $ 12.06 \pm 1.30 $
        & $ 0.362 \pm 0.001 $
        & 
        & $ 6.88 \pm 0.63 $
        & $ 92.32 \pm 0.54 $
        & $ 12.80 \pm 0.72 $
        & $ -0.154 \pm 0.004 $
        \\
        QVPO
        & $ 99.88 \pm 0.00 $
        & $ 0.04 \pm 0.00 $
        & $ 9.47 \pm 1.08 $
        & $ 0.642 \pm 0.001 $
        & 
        & $ 96.72 \pm 0.08 $
        & $ 3.20 \pm 0.08 $
        & $ 9.76 \pm 0.99 $
        & $ 0.607 \pm 0.002$
        \\
        QSM
        & $ 99.60 \pm 0.00 $
        & $ 0.40 \pm 0.00 $
        & $ \mathbf{8.40} \pm \mathbf{0.03} $
        & $ \mathbf{0.669} \pm \mathbf{0.000} $
        & 
        & $ 98.72 \pm 0.00 $
        & $ 1.20 \pm 0.00 $
        & $ \mathbf{8.62} \pm \mathbf{0.07} $
        & $ \mathbf{0.654} \pm \mathbf{0.000}  $
        \\
        QSM-A (w/ annealing)
        & $ \mathbf{99.92} \pm \mathbf{0.00} $ 
        & $ \mathbf{0.00} \pm \mathbf{0.00} $ 
        & $ 8.72 \pm 0.38 $ 
        & $ 0.663 \pm 0.000 $ 
        & & $ \mathbf{98.84} \pm \mathbf{0.00} $ 
        & $ \mathbf{0.92} \pm \mathbf{0.00} $ 
        & $ 8.91 \pm 0.37$ 
        & $ 0.649 \pm 0.000$ 
        \\ 
        \bottomrule
    \end{tabular}
    }
    \vspace{-4mm}
\end{table*}\par
\begin{algorithm}[b]
    \SetAlgoLined
    \DontPrintSemicolon
    \caption{Action generation with guidance for static obstacle avoidance}
    \label{alg:guidance}
    {\footnotesize
    \KwIn {Trained score network $\Psi_\theta$, states $\boldsymbol{s}_t = \left[\boldsymbol{s}_t^r, \boldsymbol{s}_t^h\right]$ at time $t$, 
    current robot position $\boldsymbol{p}^r_{t}$ at time $t$, robot radius $d_{r}$, nearest point of the wall $\boldsymbol{p}^w$, number of denoising steps $T$, parameters for noise scheduling $\alpha_\tau$, $\bar\alpha_\tau$, $\beta_\tau$, and $\bar\beta_\tau$} 
    $\boldsymbol{x}_T \sim \mathcal{N}(\bm{0}, \bm{I})$ \;

    \For{$\tau=T$ \KwTo $1$}{

        Denoise by score and guidance employing {

$\hat{\boldsymbol{x}}_0=\frac{1}{\sqrt{\bar\alpha_\tau}}\boldsymbol{x}_T-\frac{\sqrt{1-\bar\alpha_\tau}}{\sqrt{\bar\alpha_\tau}}\Psi_\theta(\boldsymbol{s}_t,\boldsymbol{x}_\tau,\tau)$\;
$\sigma_\tau= \sqrt{\frac{1-\bar{\alpha}_{\tau-1}}{\,1-\bar{\alpha}_\tau\,}\,\beta_\tau}$\;
        \textcolor{black}{Compute guidance cost $\mathcal{J}_w(\hat{\boldsymbol{x}}_0)$:}
        \(
\begin{aligned}
\boldsymbol{v}_{\text{before}} &= \boldsymbol{p}^w - \boldsymbol{p}_t^r, \quad
\boldsymbol{v}_{\text{after}}  = \boldsymbol{p}^w - (\boldsymbol{p}_t^r+\Delta t\hat{\boldsymbol{x}}_0) \\
c &= \frac{\boldsymbol{v}_{\text{before}} \cdot \boldsymbol{v}_{\text{after}}}{\|\boldsymbol{v}_{\text{before}}\|\|\boldsymbol{v}_{\text{after}}\| + \epsilon} \\
\xi &= \|\boldsymbol{v}_{\text{after}} \|- d_r \\
\mathcal{J}_w(\hat{\boldsymbol{x}}_0) &= 
\begin{cases}
    0, & \boldsymbol{v}_{\text{before}} \cdot \hat{\boldsymbol{x}}_0 < 0 \And  c > 0 \\
    c \cdot \exp\!\left(- \xi^2\right), & \boldsymbol{v}_{\text{before}} \cdot \hat{\boldsymbol{x}}_0 \ge 0 \And  c > 0 \\[6pt]
    -c\cdot (\lvert \xi \rvert+1), & c \leq 0 \\
\end{cases}
\end{aligned}
\)
\textcolor{black}{$\hat{\boldsymbol{x}}_0 \;\leftarrow\; \hat{\boldsymbol{x}}_0 - \sigma_\tau\nabla_{\boldsymbol{x}_\tau} \mathcal{J}_w(\hat{\boldsymbol{x}}_0)$\;}
    $\boldsymbol{\mu}_\tau = \frac{\beta_\tau \sqrt{\bar{\alpha}_{\tau-1}}}{\,1-\bar{\alpha}_\tau\,}\,\hat{\boldsymbol{x}}_0
\;+\;\frac{\sqrt{\alpha_\tau}\,\bigl(1-\bar{\alpha}_{\tau-1}\bigr)}{\,1-\bar{\alpha}_\tau\,}\,\boldsymbol{x}_\tau$\;

     $\boldsymbol{x}_{\tau-1}\sim \mathcal{N}(\boldsymbol{\mu}_\tau,\sigma_\tau\boldsymbol{I})$\;
    }
        } 
    Execute action $\boldsymbol{a}_t = \boldsymbol{x}_0$\; 
    }
\end{algorithm}
To address this, we incorporate guidance into the diffusion model to steer actions away from static obstacles, allowing the robot to avoid both pedestrians and static obstacles. In particular, the guidance strength increases as the robot approaches static obstacles, directing its movement away from the obstacles.
The execution procedure for this guidance is shown in Algorithm \ref{alg:guidance}. 

\textcolor{black}{
In this algorithm, $\mathcal{J}_w(\hat{\boldsymbol{x}}_0)$ is calculated using the distance from the nearest point $p^w$ of the static obstacles before the robot moves according to the pseudo-noise-removed action $\hat{\boldsymbol{x}}_0$ inferred in each generation step $\tau$.
In addition, by utilizing the dot product of the vector from the robot's pre-movement position to $p^w$ and $\hat{\boldsymbol{x}}_0$, and the cosine similarity $c$ of the vectors from the robot's pre- and post-movement positions to $p^w$, case distinctions are made; if the robot approaches static obstacles after moving, a gradual exponential cost is applied, but if the robot crosses over $p^w$ after taking action, a linearly increasing cost with an offset is applied to impose a larger penalty. If the robot neither approaches nor crosses static obstacles, the cost is 0.
By guiding action generation using the gradient of $\mathcal{J}_w(\hat{\boldsymbol{x}}_0)$, it is possible to avoid static obstacles in addition to pedestrians.
}
\subsection{Guidance for Companion Tasks}
As a second example of guidance, we describe a case of companion tasks in which the robot aims to reach its destination while accompanying target pedestrians. 
During training, our method only targets avoiding pedestrians to reach the goal. Therefore, without guidance, it does not generate actions specifically designed to accompany them. The guidance strategy utilizes the target pedestrian's previous position information to generate actions that minimize the distance between the robot and the previous position of the target pedestrian. This enables action generation in which agents avoid other pedestrians while accompanying the target pedestrian toward their goal. The execution procedure for this guidance is presented in Algorithm \ref{alg:guidance_ac}.
In this algorithm, $\mathcal{J}_c(\hat{\boldsymbol{x}}_0)$ is calculated as the distance between the position the robot moves to, based on the pseudo-noise-removed action $\hat{\boldsymbol{x}}_0$ estimated at each generation step $\tau$, and the position of the pedestrian from the previous step.
The companion task is achieved by guiding action generation using the gradient of $\mathcal{J}_c(\hat{\boldsymbol{x}}_0)$.
\begin{algorithm}[t]
    \SetAlgoLined
    \DontPrintSemicolon
    \caption{Action generation with guidance for companion tasks}
    \label{alg:guidance_ac}
    {\footnotesize
    \KwIn {Trained score network $\Psi_\theta$, states $\boldsymbol{s}_t = \left[\boldsymbol{s}_t^r, \boldsymbol{s}_t^h\right]$ at time $t$, 
    current robot position $\boldsymbol{p}^r_{t}$, previous position of the target pedestrian $\boldsymbol{p}^i_{t-1}$, number of denoising steps $T$, parameters for noise scheduling $\alpha_\tau$, $\bar\alpha_\tau$, $\beta_\tau$, and $\bar\beta_\tau$} 
    $\boldsymbol{x}_T \sim \mathcal{N}(\bm{0} , \bm{I})$ \;
    \For{$\tau=T$ \KwTo $1$}{
        Denoise by score and guidance employing {

$\hat{\boldsymbol{x}}_0=\frac{1}{\sqrt{\bar\alpha_\tau}}\boldsymbol{x}_T-\frac{\sqrt{1-\bar\alpha_\tau}}{\sqrt{\bar\alpha_\tau}}\Psi_\theta(\boldsymbol{s}_t,\boldsymbol{x}_\tau,\tau)$\;
$\sigma_\tau= \sqrt{\frac{1-\bar{\alpha}_{\tau-1}}{\,1-\bar{\alpha}_\tau\,}\,\beta_\tau}$\;
\textcolor{black}{Compute guidance cost $\mathcal{J}_c(\hat{\boldsymbol{x}}_0)$:}\;
    \textcolor{black}{$\mathcal{J}_c(\hat{\boldsymbol{x}}_0)=||\boldsymbol{p}^i_{t}-(\boldsymbol{p}^r_{t}+\Delta t\boldsymbol{\hat{x}}_0)||^2$\;}
\textcolor{black}{$\hat{\boldsymbol{x}}_0 = \hat{\boldsymbol{x}}_0 - \sigma_\tau\nabla_{\boldsymbol{x}_\tau} \mathcal{J}_c(\hat{\boldsymbol{x}}_0)$\;}
    $\boldsymbol{\mu}_\tau=\frac{\beta_\tau \sqrt{\bar{\alpha}_{\tau-1}}}{\,1-\bar{\alpha}_\tau\,}\,\boldsymbol{x}_0
\;+\;\frac{\sqrt{\alpha_\tau}\,\bigl(1-\bar{\alpha}_{\tau-1}\bigr)}{\,1-\bar{\alpha}_\tau\,}\,\boldsymbol{x}_\tau$\;
     $\boldsymbol{x}_{\tau-1}\sim \mathcal{N}(\boldsymbol{\mu}_\tau,\sigma_\tau\boldsymbol{I})$\;
    }
        } 
    Execute the action $\boldsymbol{a}_t = \boldsymbol{x}_0$\; 
    }
\end{algorithm}
\section{EXPERIMENTS}
Experiments were conducted in a simulation environment to evaluate the effectiveness of the proposed method.
\subsection{Simulation Environment and Settings}
We employed the CrowdNav environment adopted in some related studies \cite{Chen2019-ju, Chen2020-rj}. 
In this environment, pedestrians are controlled using ORCA \cite{VanDenBerg2011}. 
The models were trained for 100,000 episodes in a visible circle crossing scenario with five pedestrians. In addition, data were collected using ORCA before each training session, and the training began with a dataset of 2000 episodes. \textcolor{black}{The hyperparameter $\alpha$ of QSM and $\hat{\alpha}$ of QSM-A (QSM with annealing) were set to 400. The number of generation steps was set to 100 for each diffusion-based reinforcement learning method.
}
\begin{figure*}[ht]
 \centering
  \vspace{2mm}
  \includegraphics[trim={0 0 0 0}, clip, width=155mm]{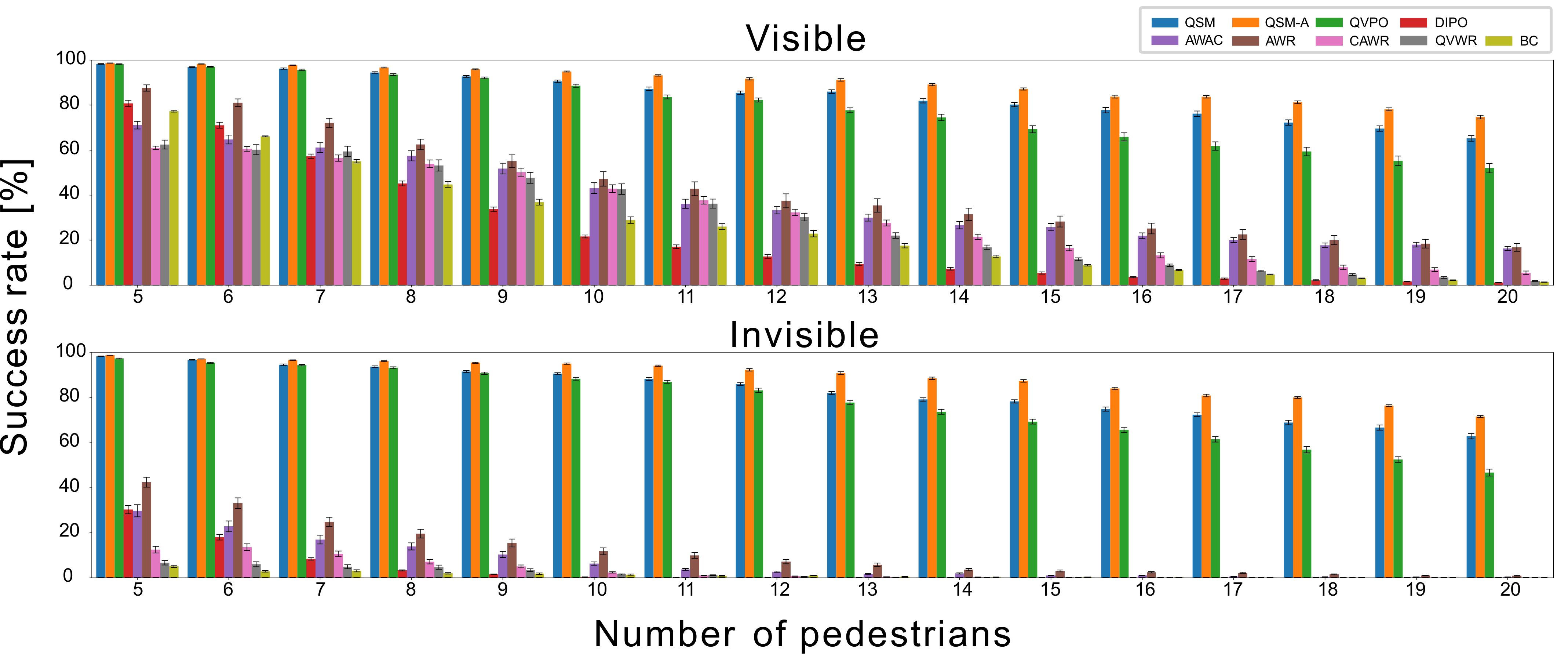}
  \caption{Comparison of success rate while changing the number of pedestrians. The upper graph shows the results for the visible setting, while the lower graph presents the results for the invisible setting.}
  \label{fig:result_success_time}
  \vspace{-3mm}
\end{figure*}
\subsection{\textcolor{black}{Performance Evaluation}}
The performance of each method was evaluated in a circle crossing scenario with pedestrians controlled by ORCA and another scenario with pedestrians controlled by the social force model. 
In each scenario, the methods were evaluated under two conditions: one where pedestrians were aware of the robot’s position (visible) and another where they were unaware (invisible). 
For comparison, the proposed method was evaluated against BC, CAWR \cite{Zhang2021-uy}, which tackles social navigation with continuous action spaces, and related reinforcement learning methods, such as AWR \cite{Peng2019-ef}, QVWR \cite{Kozakowski2022-wr}, and AWAC \cite{Nair2020-te}. Furthermore, comparisons were made with other diffusion-based reinforcement learning methods, including DIPO \cite{Yang2023-te} and QVPO \cite{Ding2024-dq}.
For each method, five models were trained using different random seeds.
The evaluation metrics included success rate, collision rate, average execution time, and average return. The test scenarios comprised 500 episodes each.\par
\textcolor{black}{
The results are summarized in Tables \ref{tab:orca_circle} and \ref{tab:socialforce_circle}. The $\pm$ symbol in each table indicates the standard deviation computed from the results of training runs using five independent random seeds. 
The tables demonstrate that QSM-based methods achieve superior performance in both ORCA and social force model \cite{Helbing1995-sm} scenarios with visible and invisible settings. While QSM demonstrates slightly faster average execution time and superior performance from the aspect of return in comparison between QSM and QSM-A, QSM-A exhibits a consistent tendency to achieve the highest success rate and the lowest collision rate.
}
Both Tables \ref{tab:orca_circle} and \ref{tab:socialforce_circle} demonstrate that methods other than QSM-based approaches show significant performance degradation in the invisible setting.
These results suggest that the QSM-based method outperforms other approaches, maintaining high performance even when pedestrian behavior patterns change, such as in the invisible setting or when using the social force model for pedestrian simulation in the environment.
\par
Fig. \ref{fig:result_success_time} shows a comparison of success rates with an increasing number of pedestrians in the circle crossing scenario under both visible and invisible settings. 
As shown in the figure, QSM-A consistently outperformed the other approaches under all conditions of the number of pedestrians.
The performance gap with QVPO is small for five pedestrians but widens as the number of pedestrians increases. This trend may be attributed to QVPO's reliance on weighted regression, which limits generalization to unseen scenarios.
Compared with other Gaussian policy-based methods, such as AWR, CAWR, QVWR, and AWAC, the proposed QSM-based approach demonstrates significantly superior performance. This indicates that the QSM-based method outperforms Gaussian policy-based approaches, maintaining high performance even when the number of pedestrians changes in the environment.

\subsection{Evaluation of Guidance for Static Obstacle Avoidance}
\begin{figure}[h]
    \centering
    \includegraphics[trim={0 5mm 0 7mm}, clip, width=70mm]{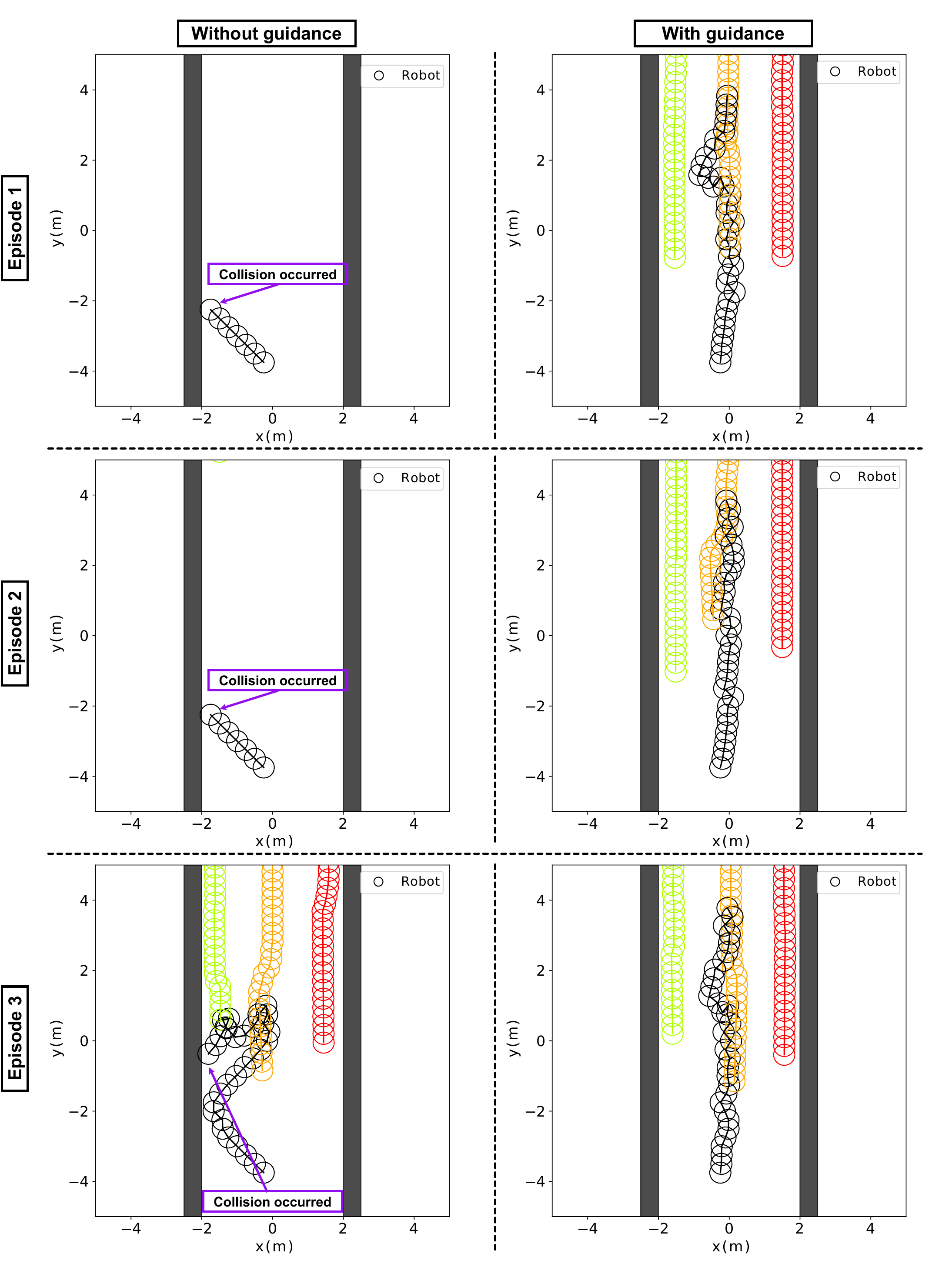}
    \caption{Comparison between with and without guidance for static obstacle avoidance. Colored circles represent pedestrian trajectories, and the black rectangles represent walls.}
    \label{fig:wall_guidance}
    \vspace{-3mm}
\end{figure}

The effectiveness of the guidance for static obstacle avoidance was evaluated in an environment with walls and three pedestrians approaching from the front. Fig. \ref{fig:wall_guidance} presents a qualitative comparison. 
The results show that when the robot initially intended to move left to avoid pedestrians, leading to a potential collision with the wall, shifting to the right by the guidance instead allowed it to avoid both pedestrians and the wall.
Table \ref{tab:wall_guidance} presents a quantitative comparison of cases with and without guidance. In the table, Col-P represents the collision rate with pedestrians, and Col-W represents the collision rate with walls. The results show that the proposed guidance method significantly reduces collisions with walls and greatly improves the success rate. 
Although the collision rate with pedestrians increases compared with the case without guidance, this likely occurs because the robot collides with walls before encountering pedestrians in the case without guidance.
This phenomenon can be attributed to the slight increase in the performance of execution time observed in the absence of guidance compared to the presence of guidance.
These results demonstrate that the proposed method can guide action generation to enable both pedestrian avoidance and navigation around static obstacles not considered during the training phase.

\begin{table}[h]
    \centering
    \caption{\textcolor{black}{Numerical comparison in environments with static obstacles}}
    \label{tab:wall_guidance}
    \scalebox{0.75}{
    \begin{tabular}{lccccc}
        \toprule
        Method & Success [\%]$\uparrow$ & Col-P [\%]$\downarrow$ & Col-W [\%]$\downarrow$ & Exec. time [s]$\downarrow$ & Return$\uparrow$ \\ \hline
        w/o guidance  & $ 0.6 $ & $ \mathbf{0.00} $ & $ 99.4 $ & $ \mathbf{8.17} $ & $-0.226$ \\ 
        w/ guidance & $ \mathbf{97.2} $ & $ 2.8 $ & $ \mathbf{0.0} $ & $ 8.46 $ & $\mathbf{0.640}$ \\ 
        \bottomrule
    \end{tabular}
    }
    \vspace{-1mm}
\end{table}\par
\subsection{Evaluation of Guidance for Companion Tasks}
\begin{figure}[h]
    \centering
    \includegraphics[trim={0 5mm 0 7mm}, clip, width=70mm]{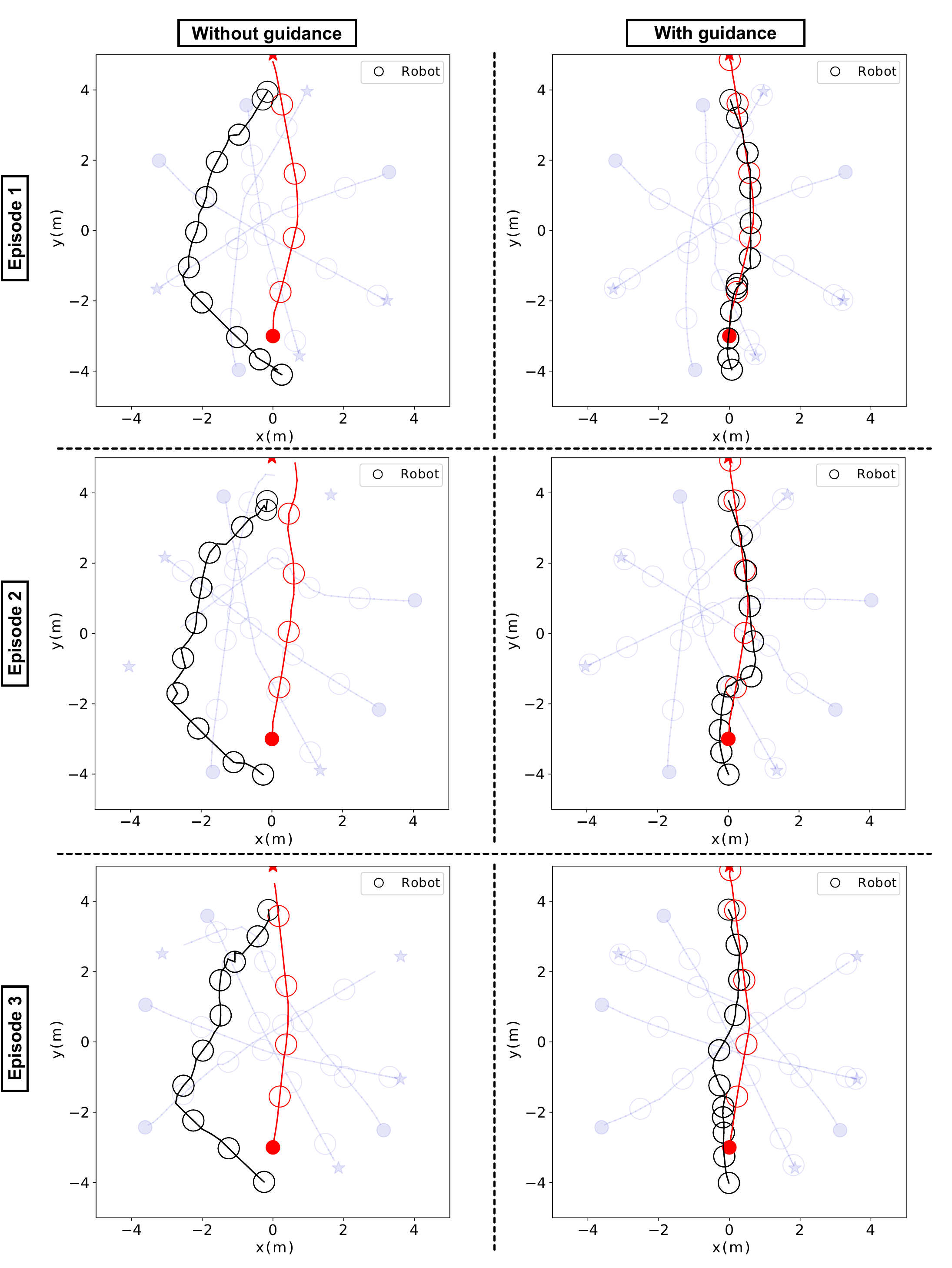}
    \caption{\textcolor{black}{Comparison between with and without guidance for companion tasks. The red circles indicate target pedestrians trajectories, and blue trajectories indicate other pedestrians.}}
    \label{fig:accompany_task}
    \vspace{-5mm}
\end{figure}

\textcolor{black}{
The effectiveness of the guidance for companion tasks was evaluated in a scenario involving five pedestrians. 
The assessment focused on two key aspects: qualitative evaluation of the robot's behavioral changes in response to the guidance in directing the target pedestrian, and a performance evaluation to ensure that the introduction of this guidance system did not result in significant performance degradation. 
Fig. \ref{fig:accompany_task} presents a qualitative comparison. These results demonstrate that when using guidance, the robot follows pedestrian movements more closely along trajectories than when guidance is not used, confirming that the guidance system successfully generates actions that accompany pedestrians.
The results of quantitative evaluation are presented in Table \ref{tab:accompany}. In the table, FD denotes the discrete Fr\'{e}chet distance between the robot and the target pedestrian. The result shows that when using guidance, the robot follows pedestrian movements more closely along trajectories compared to when guidance is not used, confirming that the guidance system successfully generates actions that accompany pedestrians.
This result demonstrates that while overall performance decreases when using the guidance system, success rates remain above 90\%, and the Fr\'{e}chet distance is shortened by approximately 40\%. 
From these results, the proposed guidance method can modify robot behavior without causing significant performance degradation, even for tasks containing objectives not considered during training.
}
\begin{table}[htpb]
    \centering
    \caption{\textcolor{black}{Numerical comparison for companion tasks}}
    \label{tab:accompany}
    \scalebox{0.75}{
    \begin{tabular}{lccccc}
        \toprule
        Method & Success [\%]$\uparrow$ & Collision [\%]$\downarrow$  & Exec. time [s]$\downarrow$ & Return$\uparrow$ & \textcolor{black}{FD [m]$\downarrow$} \\ \hline
        w/o guidance  & $ \mathbf{96.4} $ & $ \mathbf{3.6} $ &  $ \mathbf{9.09} $ & $\mathbf{0.619}$ & $2.07$ \\ 
        w/ guidance & $ 92.0 $ & $ 7.4 $ & $ 10.48 $ & $0.549$ & $\mathbf{1.20}$ \\ 
        \bottomrule
    \end{tabular}
    }
    \vspace{-2mm}
\end{table}\par
\section{Real-World Demonstration}
\textcolor{black}{
A real-world demonstration was conducted using the developed robot system across three scenarios. This demonstration verifies the applicability of our proposed method to actual robotic systems. The system is based on a Mecanum rover and is equipped with a 2D-LiDAR (UST-20LX) and onboard computer (Jetson AGX Orin) for processing. The software system was built using ROS 2 Humble, and pedestrian detection was performed using LFE \cite{Amodeo2025-hz}. 
In addition, localization was realized using adaptive Monte Carlo localization (AMCL).
For processing pedestrian detection and the proposed method, we utilized a computer equipped with an AMD Ryzen 9 7900 CPU and NVIDIA GeForce RTX 4090 GPU to distribute computational load.
}
\subsection{Circle Crossing Scenario}
\begin{figure}[htpb]
    \centering
    \vspace{-2mm}
    \begin{overpic}[width=60mm]{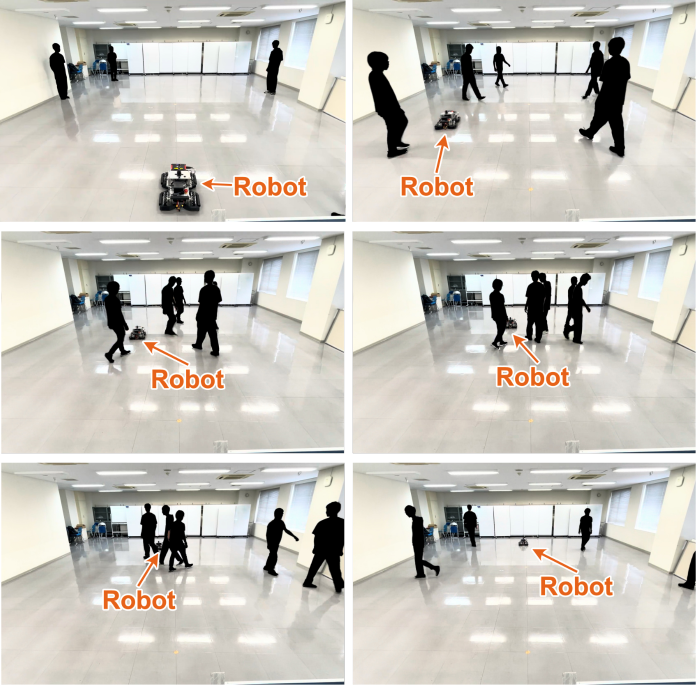}
        \put(1.5,93.0){\Circled[fill color=white]{1}}
        \put(51.5,93.0){\Circled[fill color=white]{2}}
        \put(1.5,59.0){\Circled[fill color=white]{3}}
        \put(51.5,59.0){\Circled[fill color=white]{4}}
        \put(1.5,26.0){\Circled[fill color=white]{5}}
        \put(51.5,26.0){\Circled[fill color=white]{6}}
    \end{overpic}
    \centering
    \caption{Scenes of the real-world demonstration using the proposed method in the circle crossing scenario.}
    \label{fig:real_world_ex_scene}
    \vspace{-3mm}
\end{figure}

\textcolor{black}{
We verified whether an actual robot could successfully execute the same circle-crossing scenario used in the simulation evaluation. Fig. \ref{fig:real_world_ex_scene} shows the actual demonstration of the robot navigating through pedestrians while reaching its destination. This confirms that the proposed method can be effectively applied even when operating a physical robot in real-world environments.
}
\subsection{Corridor Scenario}
\begin{figure}[htpb]
    \centering
    \vspace{2mm}
    \begin{overpic}[width=62mm]{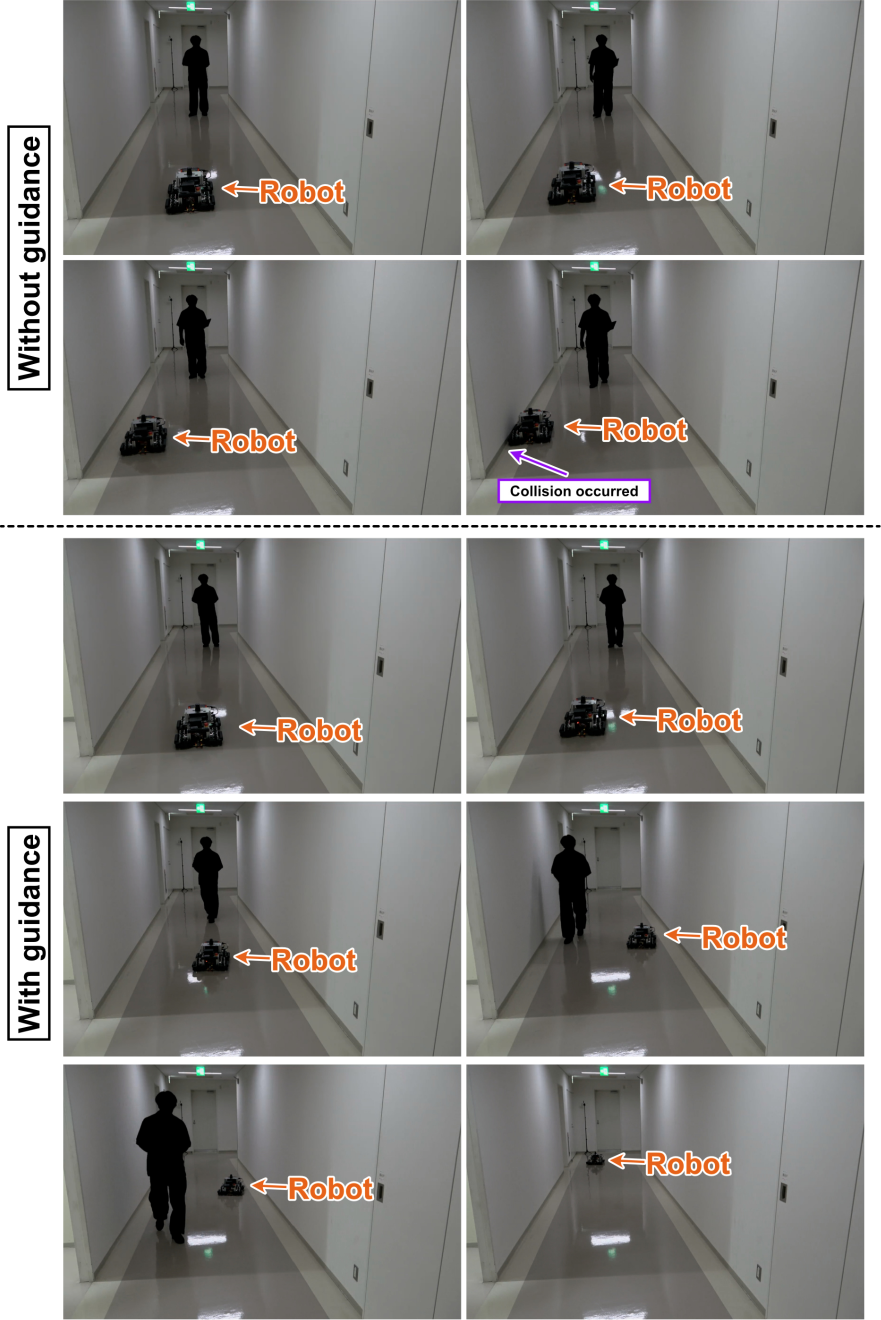}
        \put(5.8,96.0){\Circled[fill color=white]{1}}
        \put(36.5,96.0){\Circled[fill color=white]{2}}
        \put(5.8,76.5){\Circled[fill color=white]{3}}
        \put(36.5,76.5){\Circled[fill color=white]{4}}

        \put(5.8,55.5){\Circled[fill color=white]{1}}
        \put(36.5,55.5){\Circled[fill color=white]{2}}
        \put(5.8,35.0){\Circled[fill color=white]{3}}
        \put(36.5,35.0){\Circled[fill color=white]{4}}
        \put(5.8,15.5){\Circled[fill color=white]{5}}
        \put(36.5,15.5){\Circled[fill color=white]{6}}
    \end{overpic}
    \centering
    \caption{\textcolor{black}{Scenes of the real-world demonstration using the proposed method in the corridor scenario.}}
    \label{fig:real_world_wall_avoidance}
    \vspace{-1mm}
\end{figure}

\textcolor{black}{
We experimentally verified whether the guidance method for static obstacle avoidance could be successfully implemented on an actual robot in a corridor environment. The width of the corridor is approximately 2m, and there was one pedestrian in the environment that had to be avoided.
Fig. \ref{fig:real_world_wall_avoidance} demonstrates the results without guidance and with guidance.
Without guidance, the robot collided with the wall. In contrast, with guidance, the robot successfully avoided both pedestrians and static obstacles. These results confirm that the proposed guidance method for static obstacle avoidance can effectively guide action generation not only for pedestrians but also for static obstacles when deploying a physical robot in real-world environments.
}
\subsection{Companion Scenario}
\begin{figure}[htpb]
    \centering
    \begin{overpic}[width=62mm]{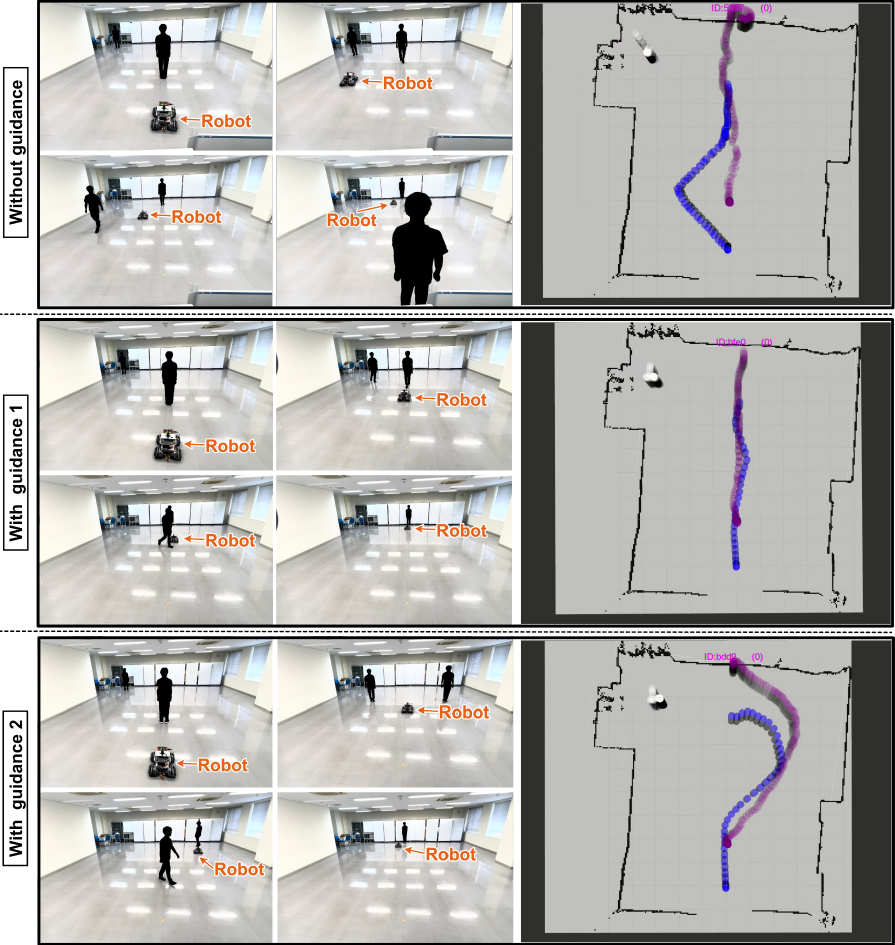}
        \put(5.0,95.0){\Circled[fill color=white]{1}}
        \put(30.0,95.0){\Circled[fill color=white]{2}}
        \put(5.0,78.0){\Circled[fill color=white]{3}}
        \put(30.0,78.0){\Circled[fill color=white]{4}}
        
        \put(5.0,61.0){\Circled[fill color=white]{1}}
        \put(30.0,61.0){\Circled[fill color=white]{2}}
        \put(5.0,45.0){\Circled[fill color=white]{3}}
        \put(30.0,45.0){\Circled[fill color=white]{4}}
         
        \put(5.0,28.0){\Circled[fill color=white]{1}}
        \put(30.0,28.0){\Circled[fill color=white]{2}}
        \put(5.0,11.0){\Circled[fill color=white]{3}}
        \put(30.0,11.0){\Circled[fill color=white]{4}}
       
    \end{overpic}
    \centering
    \caption{\textcolor{black}{Scenes of the real-world demonstration using the proposed method in the companion scenario. In the images on the right of each row, the blue markers indicate trajectories of the robot and the purple markers indicate trajectories of the target pedestrian.}}
    \label{fig:real_world_accompany}
    \vspace{-5mm}
\end{figure}

\textcolor{black}{
We evaluated whether the guidance system for companion tasks could successfully generate appropriate actions in a real-world companion scenario. Two pedestrians are placed in the environment, one is the target pedestrian and the other is the pedestrian to be avoided. Fig. \ref{fig:real_world_accompany} shows the demonstration results. When operating without guidance, the robot simply avoids pedestrians, while with guidance, it successfully generates behavior that accompanies the target pedestrian. Notably, in the results of guidance 2, the pedestrian significantly moved toward the right side of the environment, with the corresponding robot also exhibiting substantial changes in its behavior. These results confirm that our guidance system for companion tasks can successfully direct behavior generation in an actual robot operating in an actual environment, effectively guiding the robot to accompany specific pedestrians.
}
\section{CONCLUSIONS}
\textcolor{black}{
In this study, we applied diffusion-based reinforcement learning trained with the QSM-based method for social navigation and demonstrated that it outperforms other methods. In addition, we demonstrated that by utilizing guidance, conditions that were not considered during training could be incorporated without additional training.
In future work, we plan to extend the applicability of the proposed method to a wider range of scenarios and conduct more detailed real-world experiments to further enhance its performance. 
}









\bibliographystyle{ieeetr}
\bibliography{ref}

\end{document}